\pdfoutput=1
\documentclass[11pt]{article}
\PassOptionsToPackage{table}{xcolor}
\usepackage[final]{coling}
\usepackage{times}
\usepackage{latexsym}
\usepackage[T1]{fontenc}
\usepackage[utf8]{inputenc}
\usepackage{microtype}
\usepackage{inconsolata}
\usepackage{graphicx}
\usepackage{booktabs}
\usepackage{tcolorbox}
\usepackage{listings}
\lstdefinestyle{plain}{
    basicstyle=\fontsize{7}{9.5}\ttfamily,
    keywordstyle=\color{blue},
    commentstyle=\color{gray},
    stringstyle=\color{green},
    showstringspaces=false,
    breaklines=true,
    breakatwhitespace=false,
    breakindent=0pt,
    escapeinside={(*@}{@*)}
}

\definecolor{MutedGreen}{RGB}{85, 170, 85}
\definecolor{CoolAccent}{RGB}{120, 145, 230}
\definecolor{IAP}{RGB}{255, 126, 121}
\definecolor{CoT}{RGB}{91, 155, 213}
\definecolor{WarmOrange}{RGB}{255, 165, 85}

\usepackage{amsmath,amsfonts,bm}

\def\eqref#1{equation~\ref{#1}}

\def\1{\bm{1}}

\def\vi{{\bm{i}}}

\def\vp{{\bm{p}}}

\def\vu{{\bm{u}}}

\DeclareMathAlphabet{\mathsfit}{\encodingdefault}{\sfdefault}{m}{sl}
\SetMathAlphabet{\mathsfit}{bold}{\encodingdefault}{\sfdefault}{bx}{n}

\def\sD{{\mathbb{D}}}

\title{Detecting Conversational Mental Manipulation \\ with Intent-Aware Prompting}

\author{
    Jiayuan Ma$^\spadesuit$\thanks{Equal contribution.}, Hongbin Na$^\heartsuit$\footnotemark[1], Zimu Wang$^\clubsuit$, Yining Hua$^\triangle$,\\
    \textbf{Yue Liu$^\diamondsuit$, Wei Wang$^\clubsuit$, Ling Chen$^\heartsuit$} \\
    $^{\spadesuit}$The University of Sydney \quad
    $^{\heartsuit}$University of Technology Sydney \\
    $^{\clubsuit}$Xi'an Jiaotong-Liverpool University \quad
    $^{\triangle}$Harvard University \\
    $^{\diamondsuit}$University of New South Wales \\
    \texttt{jima3429@uni.sydney.edu.au, hongbin.na@student.uts.edu.au} \\
    \texttt{zimu.wang19@student.xjtlu.edu.cn, yininghua@g.harvard.edu}
    \\ \texttt{z5472597@ad.unsw.edu.au, wei.wang03@xjtlu.edu.cn, ling.chen@uts.edu.au}
}

\begin{document}
\maketitle

\begin{abstract}
Mental manipulation severely undermines mental wellness by covertly and negatively distorting decision-making. While there is an increasing interest in mental health care within the natural language processing community, progress in tackling manipulation remains limited due to the complexity of detecting subtle, covert tactics in conversations. In this paper, we propose Intent-Aware Prompting (IAP), a novel approach for detecting mental manipulations using large language models (LLMs), providing a deeper understanding of manipulative tactics by capturing the underlying intents of participants. Experimental results on the MentalManip dataset demonstrate superior effectiveness of IAP against other advanced prompting strategies. Notably, our approach substantially reduces false negatives, helping detect more instances of mental manipulation with minimal misjudgment of positive cases. The code of this paper is available at \url{https://github.com/Anton-Jiayuan-MA/Manip-IAP}.
\end{abstract}

\section{Introduction}

Human interactions inevitably involve varying degrees of mutual influence, from ethical persuasion based on facts to more harmful tactics like coercion and manipulation \cite{Fischer2022ThenAW}. Manipulation represents a serious concern, as it involves the deliberate control or distortion of an individual’s thoughts and emotions for personal gain \cite{Barnhill2014WhatIM}. Such manipulation can lead to detrimental mental health issues if left unchecked. The ability to detect and address these behaviors swiftly and accurately is critical for protecting individuals from potential mental health deterioration and ensuring their well-being.

Large language models (LLMs), known for their exceptional capability to process and reason over lengthy contexts \cite{peng2023doesincontextlearningfall}, are ideally suited for detecting mental manipulations. Recent studies have shown the reliability of LLMs in addressing mental health issues \cite{hua2024largelanguagemodelsmental,na-2024-cbt,na2024multisession}. One prominent research direction involves leveraging prompt engineering techniques, such as zero-shot, few-shot, chain-of-thought (CoT), and diagnosis-of-thought (DoT) prompting \cite{Chen2023EmpoweringPW,Schulhoff2024ThePR}, or fine-tuning the models on curated, annotated datasets sourced from social media platforms like Reddit and Twitter \cite{wang-etal-2024-knowledge,MentaLLaMA,qian2024domainspecific}.

To advance the analysis of manipulative dialogues, \newcite{Wang2024MentalManipAD} introduces the first dataset, MentalManip, specialized for mental manipulation detection and classification. Despite their strengths, LLMs exhibit notable difficulties in identifying manipulative dialogues; in particular, the false negative rate is almost twice the false positive rate, as evidenced by our pilot study (see Section \ref{section:Ob}). This limitation poses a substantial problem for real-world applications, where early detection of mental manipulation is critical. Building on this, \newcite{Yang2024EnhancedDO} concludes that a combination of few-shot and CoT prompting significantly enhances performance, highlighting the necessity for more advanced prompting techniques to improve LLM performance in this challenging task.

In response, we propose \textbf{Intent-Aware Prompting (IAP)}, a novel approach to enhance LLM's Theory of Mind (ToM) and its ability in detecting mental manipulations from dialogues. As shown in Figure \ref{fig:overview}, IAP leverages a distinct analysis of the underlying intents of both participants in the conversation, providing a deeper understanding of manipulative tactics. We perform extensive experiments on the MentalManip dataset, showcasing the superior effectiveness of IAP against other advanced prompting techniques, such as few-shot and CoT prompting. Notably, IAP significantly reduces the false negative rate, highlighting its practical relevance for early detection of mental manipulation in real-world applications.

Our key contributions are as follows: (1) the introduction of IAP for detecting mental manipulation in dialogues. It improves the ToM of LLMs via intent summarization, thus improving model performance on the task; (2) extensive experiments on the MentalManip dataset, which demonstrates that IAP outperforms baseline methods and substantially reduces false negatives; (3) human evaluation of the intent summarization process, confirming the high quality of the generated intents.

\begin{figure}[t!]
    \centering
    \includegraphics[width=\linewidth]{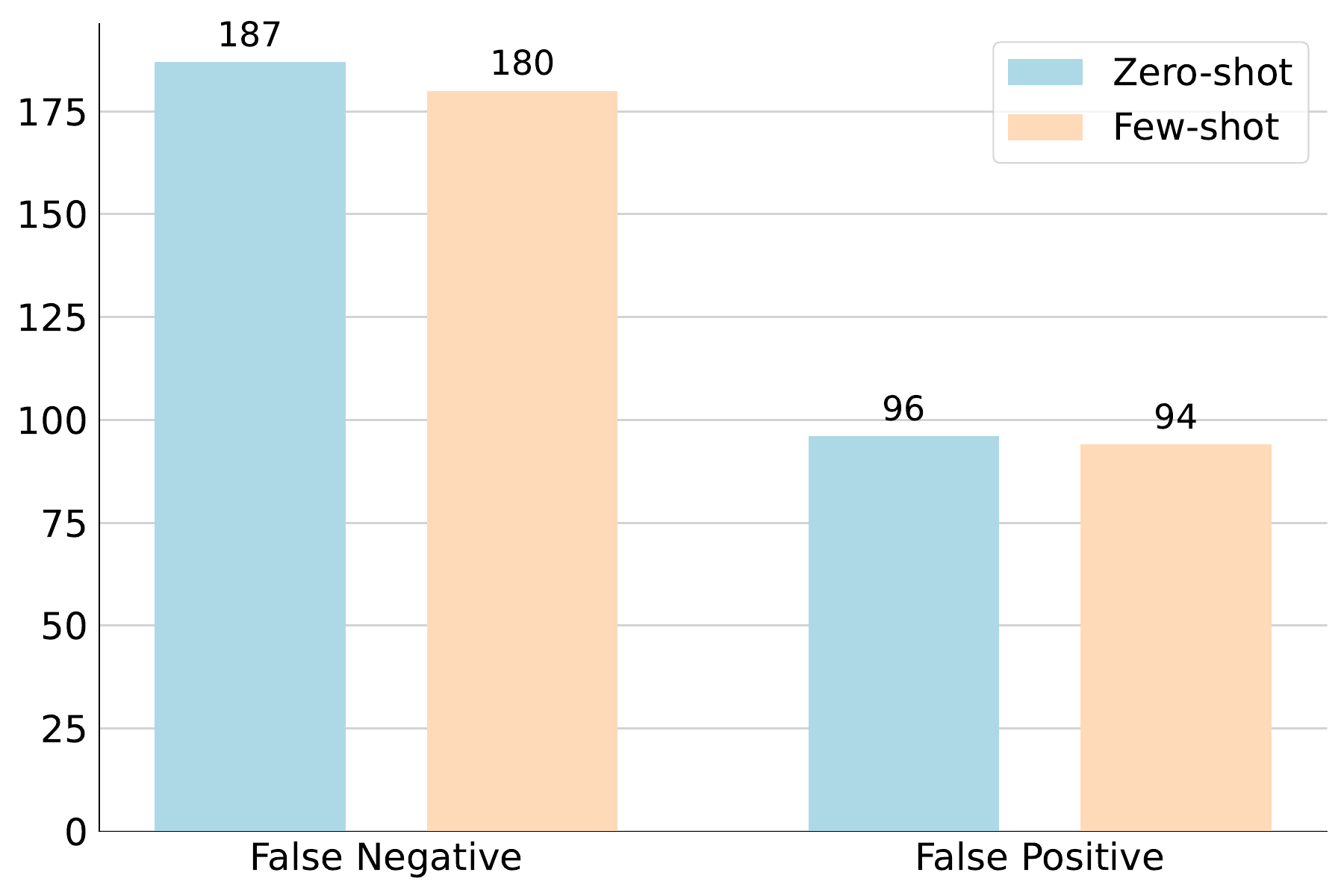}
    \caption{Comparison of false negatives and false positives in mental manipulation detection using zero-shot and few-shot prompting.}
    \label{fig:fnfp}
\end{figure}

\section{Observation}\label{section:Ob}

In pilot experiments using the zero-shot approach to detect mental manipulation \cite{Wang2024MentalManipAD}, we observe that the false negative (FN) rate is approximately \textbf{double} that of the false positive (FP) rate. This indicates challenges with the model’s ability to recognize manipulation patterns or insufficient feature representation in the input data. This observation aligns with the reality that mental manipulation is inherently difficult to detect, even for humans, due to its subtle and covert nature~\cite{Barnhill2014WhatIM}.
While \citet{Wang2024MentalManipAD} have also attempted to improve mental manipulation detection using the few-shot approach, the challenge of performance imbalance remains unresolved. The changes in FNs and FPs between the zero-shot and few-shot methods are illustrated in Figure~\ref{fig:fnfp}.

\section{Methodology}

Psychological research suggests that individuals with strong Theory of Mind (ToM) are more adept at discerning subtle differences in others' intentions \cite{byom2013theory}. Conversely, those with ToM deficits are more susceptible to manipulations \cite{kern2009theory, lampron2024profiles}. LLMs have been proven to improve ToM task performance with CoT reasoning, enhancing their ability to infer complex social cues and mental states \cite{moghaddam2023boostingtheoryofmindperformancelarge, chen-etal-2024-tombench}.

\begin{figure}[t!]
    \centering
    \includegraphics[width=0.95\linewidth]{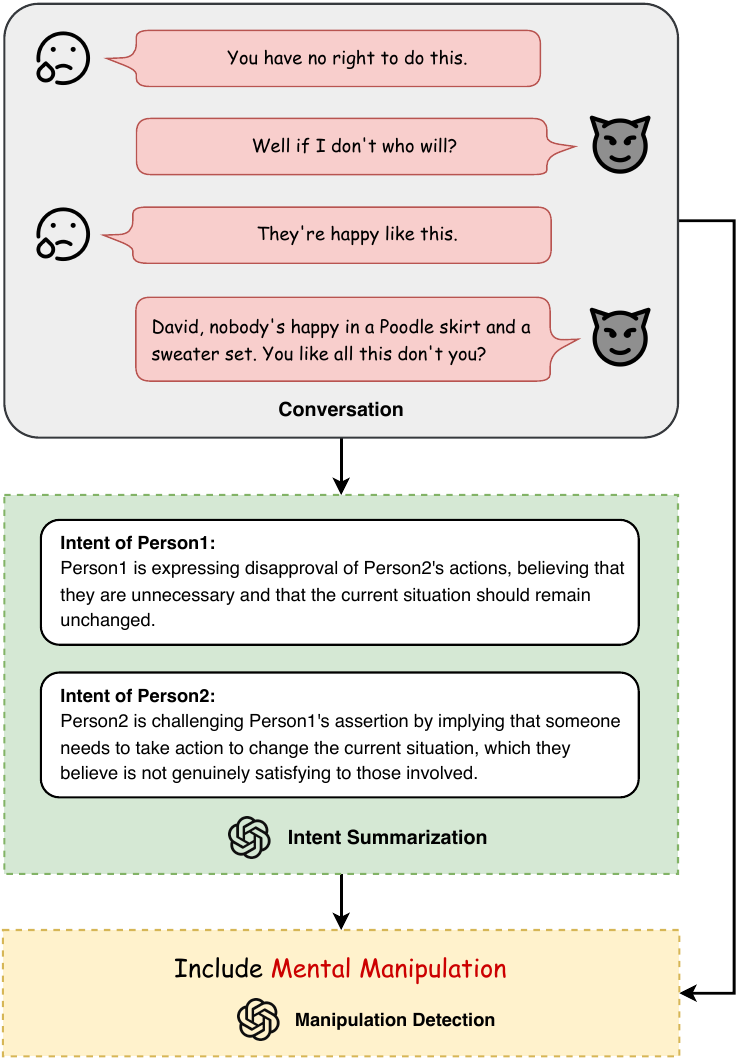}
    \caption{Overall framework of Intent-Aware Prompting (IAP) on mental manipulation detection.}
    \label{fig:overview}
\end{figure}

Building on these findings, we propose IAP for improving the ToM of LLMs and their capability in detecting mental manipulations. By incorporating intent-based reasoning, our goal is to tackle the high FN rate observed in our pilot experiments (\S\ref{section:Ob}), by improving model ability to detect subtle manipulative behavior that might be neglected. In this section, we present two key components for implementing IAP -- Intent Summarization~(\S\ref{subsect:IS}) and Manipulation Detection~(\S\ref{subsect:MD}). Figure \ref{fig:overview} shows an overview of Intent-Aware Prompting.

\begin{table*}[t!]
\centering
\resizebox{\textwidth}{!}{
\begin{tabular}{l|rrrr|rrrrrrrrrr}
\toprule
\textbf{Method} & \multicolumn{2}{c}{\textbf{FN}$\downarrow$} & \multicolumn{2}{c|}{\textbf{FP}$\downarrow$} & \multicolumn{2}{c}{\textbf{Accuracy}$\uparrow$} & \multicolumn{2}{c}{\textbf{Precision}$\uparrow$} & \multicolumn{2}{c}{\textbf{Recall}$\uparrow$} & \multicolumn{2}{c}{\textbf{F1}\textsubscript{Weighted}$\uparrow$} & \multicolumn{2}{c}{\textbf{F1}\textsubscript{Macro} $\uparrow$} \\ 
\midrule
Zero-Shot     & 187          & \multicolumn{1}{c}{-}            & 96          & \multicolumn{1}{c|}{-}         & 0.677          & \multicolumn{1}{c}{-}         & 0.813          & \multicolumn{1}{c}{-}          & 0.691          & \multicolumn{1}{c}{-}          & 0.687          & \multicolumn{1}{c}{-}         & 0.649          & \multicolumn{1}{c}{-} \\
Few-Shot      & 180          & \cellcolor[HTML]{E8F6EF}-3.7\%   & \textbf{94} & \cellcolor[HTML]{E8F6EF}-2.1\% & 0.687          & \cellcolor[HTML]{E8F6EF}1.5\% & \textbf{0.819} & \cellcolor[HTML]{E8F6EF}0.7\%  & 0.702          & \cellcolor[HTML]{E8F6EF}1.6\%  & 0.696          & \cellcolor[HTML]{E8F6EF}1.3\% & 0.659          & \cellcolor[HTML]{E8F6EF}1.5\% \\
Zero-Shot CoT & 159          & \cellcolor[HTML]{C9E9D9}-15.0\%  & 101         & \cellcolor[HTML]{FAE6E5}5.2\%  & 0.703          & \cellcolor[HTML]{E8F6EF}3.8\% & 0.815          & \cellcolor[HTML]{E8F6EF}0.2\%  & 0.737          & \cellcolor[HTML]{E8F6EF}6.7\%  & 0.710          & \cellcolor[HTML]{E8F6EF}3.3\% & 0.670          & \cellcolor[HTML]{E8F6EF}3.2\% \\
\midrule
Intent-Aware & \textbf{130} & \cellcolor[HTML]{A0D9BD}-30.5\%  & 110         & \cellcolor[HTML]{F0B3AD}14.6\% & \textbf{0.726} & \cellcolor[HTML]{E8F6EF}7.2\% & 0.812          & \cellcolor[HTML]{FAE6E5}-0.1\% & \textbf{0.785} & \cellcolor[HTML]{C9E9D9}13.6\% & \textbf{0.728} & \cellcolor[HTML]{E8F6EF}6.0\% & \textbf{0.685} & \cellcolor[HTML]{E8F6EF}5.5\% \\
\bottomrule
\end{tabular}
}
\caption{\textbf{Result of detecting mental manipulation using GPT-4.} Metrics with an upward arrow $\uparrow$ indicate higher values are better, while metrics with a downward arrow $\downarrow$ indicate lower values are better. Using zero-shot as comparison, \colorbox[HTML]{A0D9BD}{darker} \colorbox[HTML]{E8F6EF}{green} means better performance, and \colorbox[HTML]{E67C73}{darker} \colorbox[HTML]{FAE6E5}{red} means worse performance of the model.}
\label{tab:results}
\end{table*}

\subsection{Intent Summarization}\label{subsect:IS}

Consider we have a conversation $\sD$, structured as:
\begin{align*}
\sD = \{\vu_{A1}, \vu_{B1}, \dots, \vu_{An}, \vu_{Bn}\},
\end{align*}
where $u_A$ represents utterances by Person A and $u_B$ represents utterances by Person B. 
We design an intent summarization prompt $P_{\text{IS}}(\cdot)$, which consists of an intent summarization instruction for the two people $\vp_A$ and $\vp_B$. The intent summaries $\vi_A$ and $\vi_B$ can be defined as:
\begin{align}
\vi_A &= \text{LLM}(\sD, P_{\text{IS}}(\vp_A)),
\label{eq:ISA}
\\
\vi_B &= \text{LLM}(\sD, P_{\text{IS}}(\vp_B)),
\label{eq:ISB}
\end{align}
where $\text{LLM}(\cdot)$ represents an LLM used to generate the intent summary is provided in Appendix~\ref{appendix:prompt}.

\paragraph{Remark.} The entire conversation $\sD$ is used instead of just the utterances from one individual because understanding each person's intent relies on a holistic understanding of the contexts.

\subsection{Manipulation Detection} \label{subsect:MD}
Given the conversation $\sD$, and the intent summaries $\vi_A$ and $\vi_B$ calculated from $\sD$ using Equations \ref{eq:ISA} and \ref{eq:ISB}, the mental manipulation detection process can be defined as:
\begin{align}
r &= \text{LLM}(\sD, \vi_A, \vi_B, P_{\text{MD}}),
\end{align}
where $r$ denotes the detection result, with $r \in \{0, 1\}$. Specifically, $r=0$ means that no mental manipulation has been detected, while $r=1$ means that mental manipulation is present. The function $\text{LLM}(\cdot)$ refers to a LLM used to process the full conversation $\sD$ along with the intent summaries $\vi_A$ and $\vi_B$ and the manipulation detection prompt $P_{\text{MD}}$. The prompt $P_{\text{MD}}$ is specifically designed to evaluate the interaction between the two intent summaries and detect potential manipulation behaviors in the conversation. The detailed prompt is provided in Appendix~\ref{appendix:prompt}.

\section{Experiments}

\subsection{Experimental Settings}

\paragraph{Dataset.} We conducted experiments using the MentalManip dataset \cite{Wang2024MentalManipAD}, which provides multi-level annotations aimed at detecting and classifying mental manipulations. It consists of 4,000 multi-turn fictional dialogues between two characters derived from online movie scripts and includes annotation across three dimensions: the presence of manipulation, manipulation technique, and targeted vulnerability. For our experiments, we sampled a subset with 30\% instances (1.5 times of the original test set) in MentalManip$_\text{con}$, a subset of MentalManip with full annotator consensus, ensuring high quality and consistency in the experimented data.

\paragraph{Evaluation Metrics.} Following \newcite{Wang2024MentalManipAD}, we evaluated performance using accuracy, precision, recall, and F1-score (weighted and macro). Furthermore, the analysis of the false predictions, including false negatives (FN) and false positives (FP), is also crucial, as it provides insights on how well the method works to identify psychologically manipulated dialogues without excessively exaggerating the presence of the positive class.

\paragraph{Baselines.} In accordance with the previous work, we compared the performance of IAP against the following baselines: (1) \textbf{Zero-shot prompting}, which enables LLMs to perform the task based solely on the given input. (2) \textbf{Few-shot prompting} \cite{Brown2020LanguageMA}, which generalizes LLMs to the task by providing a few examples within the input prompt. We randomly selected three examples from the data subset not used for testing, with a proportion of 1:2 of manipulative to non-manipulative. (3) \textbf{Chain-of-Thought (CoT) prompting} \cite{Kojima2022LargeLM}, which enhances LLM’s reasoning capabilities by generating intermediate reasoning steps within its output, enabling more complicated problem-solving and decision-making processes.

\subsection{Experimental Results}

\paragraph{Main Results.} Table \ref{tab:results} illustrates the experimental results of IAP against baselines, in which GPT-4\footnote{\texttt{gpt-4-1106-preview}} \cite{openai2024gpt4technicalreport} was utilized in all experiments. From the table, we observed the effectiveness of IAP on mental manipulation detection by achieving the best performance on nearly all evaluation metrics, demonstrating substantial improvements in accuracy (+7.2\%), recall (+13.6\%), weighted F1-score (+6.0\%), and macro F1-score (+5.5\%) compared with zero-shot prompting. This underscores the potent efficacy of analyzing speakers' intentions in identifying the existence of mental manipulation within dialogues and the improvement of the ToM of the model. Besides, IAP achieved the lowest number of false negatives (130), representing a 30.5\% reduction compared to zero-shot prompting and substantially outperforming other baseline methods, reinforcing its ability to detect a higher number of mental manipulative dialogues. While there is a trade-off with an increase in false positives (+14.6\%), the substantial reduction in false negatives is far more critical, where early detection and intervention for potential mental health concerns are much more paramount.

\paragraph{Human Evaluation of Generated Intents.}
To assess the quality of the generated intents in the absence of references, we conducted a human evaluation to verify whether they correctly identified the manipulators. We selected 50 dialogues from the dataset that exhibited mental manipulations, and two annotators independently assessed each dialogue, labeling Person A, Person B, or both as the manipulator(s). The inter-annotator agreement reached 74\%, and the discrepancies were resolved through discussion to reach a consensus.
During evaluation, we verified if the generated intents accurately pointed to the labeled manipulator(s). As shown in Table \ref{tab:intent_evaluation}, 82\% of the intents correctly identified the manipulator(s), demonstrating the capability of IAP in producing high-quality intents that significantly aid in the detection of manipulations. Some examples are in Appendix \ref{appendix:result}.

\begin{table}[t!]
\centering
\small
\begin{tabular}{lcc}
\toprule
\textbf{Rating Category} & \textbf{Percentage} \\
\midrule
\textbf{Accurate}   & 82\% \\
\textbf{Inaccurate} & 18\% \\
\bottomrule
\end{tabular}
\caption{Percentage of intents rated as accurate and inaccurate based on human evaluation.}
\label{tab:intent_evaluation}
\end{table}

\section{Related Work}

\paragraph{Mental Manipulation Detection.}
Dialogue-based classification poses unique challenges due to the dynamic, multi-turn nature of conversations. These challenges include handling long text sequences, managing context shifts, capturing speaker roles and intents, and modeling nuanced interactions across multiple turns. In mental healthcare, dialogue-based classification has been primarily used for identifying mental health conditions \cite{ hua2024largelanguagemodelsmental} and detecting toxic behaviors \cite{ozoh2019identification}, including threats, obscenity, insults, identity-based hate, harassment, and socially disruptive persuasion \cite{toxic}.

However, the research on mental manipulation remains underexplored. The only existing work \cite{Yang2024EnhancedDO} investigates prompting techniques for mental manipulation detection with limited standard prompting techniques. Different from the previous work, we introduce a novel approach for by analyzing underlying intents of both participants in the conversation, offering a deeper understanding of manipulative tactics.

\paragraph{LLMs and Theory of Mind.}
LLMs with exceptional capability to process and reason over lengthy contexts have become the cornerstone of numerous NLP tasks \cite{peng2023doesincontextlearningfall,wang-etal-2024-document-level}, making them particularly well-suited for dialogue-based applications \cite{na2024multisession,lee-etal-2024-cactus,zheng-etal-2024-thoughts,zheng-etal-2024-self,iftikhar2024therapynlp}, which require comprehending not only individual turns but also the evolution of context, tone, and intent throughout the conversations.

``Theory of Mind'' (ToM) refers to the ability to infer and understand the mental states, intentions, beliefs, and emotions of others. While traditionally regarded a human cognitive trait, recent research suggests that LLMs can simulate aspects of this ability, even surpassing humans in tasks like recognizing irony and false beliefs \cite{strachan2024testing}. This capability is particularly valuable for mental manipulation detection, where accurately interpreting and predicting speakers' intentions and emotional states is essential for uncovering manipulative strategies in dialogues \cite{kern2009theory, lampron2024profiles}.

\section{Conclusion}

We introduced Intent-Aware Prompting (IAP), a novel approach to enhance LLM's ability in detecting mental manipulations from dialogues. It enhanced LLM's ToM and its manipulation detection capability by distinctly analyzing the underlying intents of both participants, offering a more nuanced understanding of manipulative strategies. Through comprehensive experiments on the MentalManip dataset, IAP consistently outperformed other advanced prompting techniques, such as few-shot and CoT prompting, across multiple metrics. Notably, it achieved a substantial reduction in false negatives, a crucial improvement in the context of mental health support systems where early detection of psychological manipulation is key to timely interventions. In the future, we will expand IAP to broader mental health applications to more real-world scenarios.

\section*{Limitations}

The limitations of this paper are as follows: (1) Although performance increased, the reduction in false negatives led to a slight increase in false positives. While its real-world impact is minimal, it might introduce therapeutic costs. Future research can focus on optimizing the trade-off between false negatives and false positives. (2) Due to only one dataset available, we only tested the performance of IAP on the MentalManip dataset. Future research can develop more diverse mental manipulation datasets encompassing both high-resource and low-resource languages and validate the generalizability of IAP across different linguistic and contextual settings.

\bibliography{custom}

\appendix
\clearpage
\onecolumn

\section{Prompts in Experiments}
\label{appendix:prompt}

\begin{figure}[h!]
    \begin{tcolorbox}[title=Zero-shot Prompting, left=2mm,right=1mm,top=0mm, bottom=0mm,colback=white,colframe=CoolAccent]
    \begin{lstlisting}[style=plain]
I will provide you with a dialogue. Please determine if it contains elements of mental manipulation. Just answer with 'Yes' or 'No', and don't add anything else. 

<insert dialogue>
    \end{lstlisting}
    \end{tcolorbox}

    \begin{tcolorbox}[title=Few-shot Prompting, left=2mm,right=1mm,top=0mm, bottom=0mm,colback=white,colframe=MutedGreen]
    \begin{lstlisting}[style=plain]
I will provide you with a dialogue. Please determine if it contains elements of mental manipulation. Just answer with 'Yes' or 'No', and don't add anything else. Here are 3 examples: 

Example 1:
<insert manipulative_dialogue1>
<insert manipulative_answer1>

Example 2:
<insert nonmanipulative_dialogue1>
<insert nonmanipulative_answer1>

Example 3:
<insert nonmanipulative_dialogue2>
<insert nonmanipulative_answer2>

<insert dialogue>
    \end{lstlisting}
    \end{tcolorbox}

    \begin{tcolorbox}[title=Zero-shot CoT Prompting, left=2mm,right=1mm,top=0mm, bottom=0mm,colback=white,colframe=CoT]
    \begin{lstlisting}[style=plain]
I will provide you with a dialogue. Please determine if it contains elements of mental manipulation. Just answer with 'Yes' or 'No', and don't add anything else. Let's think step by step.

<insert dialogue>
    \end{lstlisting}
    \end{tcolorbox}

    \begin{tcolorbox}[title=Intent-Aware Prompting, left=2mm,right=1mm,top=0mm, bottom=0mm,colback=white,colframe=IAP]
    \begin{lstlisting}[style=plain]
 I will provide you with a dialogue and intent of person1, and intent of person2. Please carefully analyze the dialogue and intents, and determine if it contains elements of mental manipulation. Just answer with 'Yes' or 'No', and don't add anything else.

<insert dialogue>
<insert person1's intent>
<insert person2's intent>
    \end{lstlisting}
    \end{tcolorbox}
    \caption{Examples of various prompting strategies used in the experiment: Zero-shot Prompting, Few-shot Prompting, Zero-shot CoT Prompting, and Intent-Aware Prompting. Each prompt is designed to assess the presence of mental manipulation in dialogues.}
    \label{fig:ex-direct}
\end{figure}

\clearpage
\section{Intent Summarization Examples}
\label{appendix:result}

\begin{figure}[h!]
    \begin{tcolorbox}[title=Intent Summarization Prompting, left=2mm,right=1mm,top=0mm, bottom=0mm,colback=white,colframe=WarmOrange]
    \begin{lstlisting}[style=plain]
I will provide you with a dialogue. Please summarize the intent of the statement made by <insert person> in one sentence.

<insert dialogue>
    \end{lstlisting}
    \end{tcolorbox}
    \begin{tcolorbox}[title=Intent Summarization Example 1, left=2mm,right=1mm,top=0mm, bottom=0mm,colback=white]
    \begin{lstlisting}[style=plain]
Dialogue:
Person1: You have no right to do this.
Person2: Well if I don't who will?
Person1: They're happy like this.
Person2: David, nobody's happy in a Poodle skirt and a sweater set. You like all this don't you?
```
Person1's intent:
Person1 is expressing disapproval of Person2's actions, believing that they are unnecessary and that the current situation should remain unchanged.
```
Person2's intent:
Person2 is challenging Person1's assertion by implying that someone needs to take action to change the current situation, which they believe is not genuinely satisfying to those involved.
    \end{lstlisting}
    \end{tcolorbox}

    \begin{tcolorbox}[title=Intent Summarization Example 2, left=2mm,right=1mm,top=0mm, bottom=0mm,colback=white]
    \begin{lstlisting}[style=plain]
Dialogue:
Person1: What are you doing here?
Person2: Nothing.  I just wanted you to know I was out.  I just wanted to see you.
Person1: Well, here I am.  See?
Person2: How are you doing?
Person1: George, you just can't show up, tell me you love me, and have everything be okay.
Person2: Dad.
Person1: What?
Person2: You can call me Dad if you want.
Person1: I don't want, alright?  It's not funny. I'm really pissed off, George.  You blew it, now leave me alone.
Person2: Kristina, c'mon, I'm sorry.  I'm going to make this right.  I've got a few things going on...
Person1: What do you want from me?
Person2: Just to walk with you.  I want to be your dad again.
Person1: Do what you want, it's a free country.
```
Person1's intent:
Person1 expresses frustration and anger towards Person2, indicating that Person2's past actions have caused damage to their relationship, and simply declaring love is not enough to mend it.
```
Person2's intent:
Person2 expresses a desire to reconnect and reestablish a father-daughter relationship with Person1.
    \end{lstlisting}
    \end{tcolorbox}
    \caption{Examples of intent summarization, illustrating how dialogue between two individuals can be analyzed to extract the underlying intent behind their statements. Each example provides a clear one-sentence summary for both Person1 and Person2, showcasing differing perspectives and emotional undertones within the conversations.}
\end{figure}
\end{document}